\pdfoutput=1
\documentclass[11pt]{article}

\usepackage[preprint]{acl}
\usepackage{soul}
\usepackage{times}
\usepackage{latexsym}
\definecolor{binary-color}{rgb}{0.29, 0.59, 0.82} % Blue
\definecolor{multi-class-color}{rgb}{0.9, 0.4, 0.38} % Red
\newcommand*{\WB}[1]{\framebox{#1}}
\usepackage[T1]{fontenc}
\usepackage[utf8]{inputenc}

\usepackage{microtype}

\usepackage{inconsolata}

\usepackage{graphicx}
\usepackage{times}
\usepackage{latexsym}
\usepackage{graphicx} 
\usepackage{times}
\usepackage{xcolor}
\definecolor{calpolypomonagreen}{rgb}{0.0, 0.42, 0.24}
\definecolor{carrotorange}{rgb}{0.72, 0.45, 0.2}

\definecolor{gnuplot-blue}{rgb}{0.0, 0.25, 0.58} % for Holistic
\definecolor{gnuplot-red}{rgb}{0.9, 0.4, 0.38}     % for Analytic
\newcommand{\BF}[1]{{\color{gnuplot-blue}#1}}
\newcommand{\OF}[1]{{\color{gnuplot-red}#1}}

\usepackage{latexsym}
\usepackage{array}
\usepackage{booktabs}
\usepackage{graphicx}
\usepackage{array}
\usepackage{CJKutf8}
\usepackage[utf8]{inputenc}
\usepackage{geometry}
\usepackage{xcolor}    
\usepackage{mdframed}  
\usepackage{etoolbox}
\definecolor{promptbgcolor}{rgb}{0.95,0.95,0.95}  %
\definecolor{promptlinecolor}{rgb}{0.6,0.6,0.6}  

\newcounter{prompt}

\definecolor{brightmaroon}{rgb}{0.76, 0.13, 0.28}

\usepackage{xspace}

\makeatletter
\DeclareRobustCommand\onedot{\futurelet\@let@token\@onedot}
\def\@onedot{\ifx\@let@token.\else.\null\fi\xspace}

\def\eg{\emph{e.g}\onedot}

\makeatother
\usepackage[T1]{fontenc}

\title{Contrasting Cognitive Styles in Vision-Language Models: Holistic \\ Attention in Japanese Versus Analytical Focus in English}

\author{Ahmed Sabir$^{1}$\hspace{0.04cm} 
Azinovič Gasper$^{2}$\hspace{0.04cm}
Mengsay Loem$^{3}$\thanks{This work is not associated with the author’s affiliation, Sansan, Inc.}\hspace{0.04cm}
Rajesh Sharma$^{1,4}$   \\
$^{1}$University of Tartu, Estonia, $^{2}$University of Ljubljana, Slovenia\\ 
$^{3}$Sansan, Inc., Japan, $^{4}$Plaksha University, India \\
}

\begin{document}
\maketitle
\begin{abstract}
Cross-cultural research in perception and cognition has shown that individuals from different cultural backgrounds process visual information in distinct ways. East Asians, for example, tend to adopt a holistic perspective, attending to contextual relationships, whereas Westerners often employ an analytical approach, focusing on individual objects and their attributes. In this study, we investigate whether Vision-Language Models (VLMs) trained predominantly on different languages, specifically Japanese and English, exhibit similar culturally grounded attentional patterns. Using comparative analysis of image descriptions, we examine whether these models reflect differences in holistic versus analytic tendencies. Our findings suggest that VLMs not only internalize the structural properties of language but also reproduce cultural behaviors embedded in the training data, indicating that cultural cognition may implicitly shape model outputs.
\end{abstract}

\section{Introduction}

\begin{figure*}
    \centering
    \includegraphics[width=\linewidth]{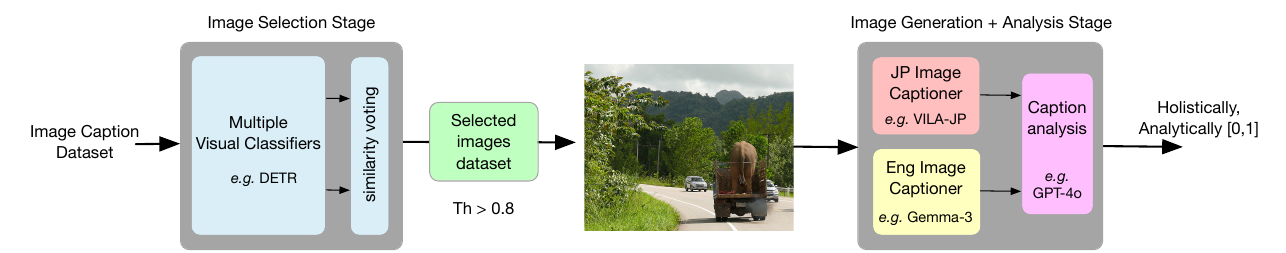}
 \vspace{-7mm}
\caption{
Overview of our analysis framework. 
Our methodology identifies whether image descriptions exhibit holistic or analytic tendencies. First, we filter and select images using four visual classifiers combined with object similarity voting. Next, we generate image captions using VLMs, \eg Gemma-3 for English and VILA-JP for Japanese. Finally, we use GPT-4o with few-shot in-context learning to classify each caption as holistic or analytic.
}
    \label{fig:enter-label}
\end{figure*}

 The influence of culture on human cognition is profound, shaping distinct patterns of thought and perception across societies \cite{Masuda2001, masuda2006culture}. For example, individuals from East Asian and North American backgrounds differ fundamentally in how they interpret visual information. These divergences are often linked to intellectual traditions: analytic thinking rooted in ancient Greek philosophy emphasizes object-focused reasoning, while East Asian traditions favor holistic thinking, highlighting the interplay between objects and their contexts \cite{masuda2017culture}.

These culturally shaped visual processing strategies extend to digital media and information foraging behaviors. Studies on mock webpages show that East Asian participants are more efficient at processing large volumes of information compared to North Americans \cite{wang2012much}. \citet{baughan2021cross} found that American users tend to rely on salient cues for rapid searching, whereas Japanese users scan content more broadly, prioritizing completeness over speed. In visual storytelling, such differences are also evident. A cross-cultural analysis of comics revealed that Japanese manga often employs minimalist or blank backgrounds, assuming contextual inference by the reader, while American and European comics depict detailed settings more explicitly \cite{atilla2023background}. These stylistic choices reflect broader cognitive expectations about the reader's attention and interpretation.

Cultural differences in visual interpretation are also mirrored in linguistic patterns. Multilingual image descriptions, such as those in Japanese and Korean, tend to encode a broader range of semantic attributes and relationships than monolingual (\eg English) descriptions \cite{ye2024computer}. Together, these findings from psychology, perception, and human-computer suggest that culturally distinct modes of visual processing are internalized and persist over time.

This raises a compelling question for the development of artificial intelligence. Given these human-like cultural divergences, do advanced systems such as Vision-Language Models (VLMs) also exhibit culturally influenced patterns of attention and description? Specifically, can models trained or fine-tuned on language-specific data, such as Japanese versus English, demonstrate holistic versus analytic tendencies akin to those seen in human observers from corresponding cultures?

This study explores whether VLMs inherit culturally situated patterns through their training data. We focus on comparing the descriptive outputs of models exposed primarily to Japanese versus English, analyzing the extent to which their outputs reflect holistic (context-sensitive) versus analytic (object-focused) styles. Our findings suggest that these models not only absorb linguistic structures but also encode cultural biases in visual interpretation, echoing attentional and descriptive tendencies observed in human cognition.

\section{Problem Statement and Definitions}
\label{sec:problem-statement}
We investigate whether VLMs trained in different languages reflect culturally influenced cognitive styles, specifically, whether Western models produce more \textit{analytic} image descriptions, while Eastern models favor \textit{holistic} descriptions. We define: \textbf{Object} as the primary subject in an image \eg cat or a bird are considered objects; \textbf{Background} as contextual elements excluding the main object \eg sky or a dining room are considered Background; Analytic description as text centered on objects; and Holistic description as text emphasizing the broader scene.

\section{Dataset}
\label{sec:dataset}

We use the COCO Caption dataset \cite{chen2015microsoft}, which contains 120K images with five human-written captions per image. It includes 82,783 training and 40,504 validation images. For our experiments, we randomly select 7K images from the validation set. A filtering process is applied to retain only images with confidently detectable objects, resulting in a subset of 5K images.

In addition, to explore cultural differences in image description more robustly, we also evaluate on: (1) HERON Benchmark \cite{inoue2024heronbench}: 21 culturally-relevant Japanese images across 7 categories (anime, art, culture, food, landscape, landmark, transportation), (2) JA-VLM \cite{akiba2024evomodelmerge}: 42 Japanese cultural images from the wild, (3) 1K Japanese Image Dataset (ours): 6435 Japanese photos filtered to 1K using human validation.\footnote{\href{https://huggingface.co/datasets/ThePioneer/japanese-photos}{https://huggingface.co/datasets/Japanese-photos}}

\begin{table*}[ht]
\centering
\small
 \resizebox{0.9\textwidth}{!}{
\begin{tabular}{l c c c c c c c c c}
\hline
 &  & \multicolumn{2}{c}{1K \%} &  & \multicolumn{2}{c}{2K \%} &  & \multicolumn{2}{c}{5K \%} \\
\addlinespace[1mm]
\cline{3-4} \cline{6-7} \cline{9-10}
\addlinespace[1mm]
Model & Method & Holistic & Analytic & & Holistic & Analytic & & Holistic & Analytic \\
\hline

\addlinespace[1mm]

GPT4o  \cite{openai2024gpt4o} & 5-shot &  22.60 & 77.40  && 20.25 & 79.75 & & 19.22 & \textbf{80.78} \\ 
GPT4o (balanced)  & 6-shot &  20.80 & 79.20  && 21.05 & 78.95 & & 19.80 & 80.20 \\                                             

\addlinespace[1mm]
\hline
\addlinespace[1mm]

GPT4o-JP   & 5-shot &   24.60 & 75.40  &&  24.15 & 75.85 & &  23.64 & 76.36 \\ 
GPT4o-JP (balanced)  & 6-shot &  26.10 & 73.90  && 27.75 & 72.25 & & \textbf{28.64} & 71.36 \\ 
\addlinespace[1mm]
\hline
\addlinespace[1mm]

Gemma-3-12B \cite{team2025gemma}  & 5-shot & 13.40 & 86.60  && 13.30 & 86.70 & & 14.16 & 85.84 \\ 
Gemma-3-12B (balanced)  & 6-shot &  13.70 & 86.30  && 13.15 & 86.85 & & 13.78 & 86.22 \\                                             
\addlinespace[1mm]
\hline
\addlinespace[1mm]
VILA-JP-14B  \cite{sasagawa2024constructing}  &  5-shot    &   11.60 & 88.40  &&  12.80 & 87.20 & &  13.88  &  86.12  \\ 
VILA-JP-14B (balanced)  &  6-shot    &   14.70 & 85.30  && 15.75  & 84.25 & &  16.90 &  83.10   \\  \addlinespace[1mm]
 \hline
\addlinespace[1mm]
Gemma-3-27B-JP  & 5-shot & 13.00 & 87.00  && 13.00 & 87.00 & & 13.50  & \textbf{\underline{86.50}} \\
Gemma-3-27B-JP (balanced)  & 6-shot &  19.70 & 80.30  && 19.55 & 80.45 & & \textbf{\underline{21.14}} & 78.86 \\    
%GPT4o   &  CoT   &  21.30 & 78.70  && 23.45 & 76.55 & & 23.82 & 76.18 \\ 
\hline

%\addlinespace[1mm]
%\hline
%\addlinespace[1mm]
%\hline

\end{tabular}}
\vspace{-0.2cm}
\caption{Comparison analysis results of different English and Japanese Vision-LLMs on a subset of the COCO dataset. GPT-4o serves as the evaluator, judging whether each model's image caption begins with a background-focused (\textit{Holistic}) or object-focused (\textit{Analytic}) description. The results indicate that larger models, such as GPT-4o, better replicate cultural cognitive styles. \textbf{Bold} indicates the highest score among large models, while \underline{\textbf{Underline}} indicates the highest score among smaller models.}
\label{tab:main-table_paper}
\end{table*}

\begin{table*}[ht]
\centering
\small
\begin{tabular}{lcccccccc}
\hline
 &  & \multicolumn{2}{c}{HERON-BENCH \%} &  & \multicolumn{2}{c}{JA-VLM \%} & \multicolumn{2}{c}{1K- JP-dataset  \%} \\
\addlinespace[1mm]
\cline{3-4} \cline{6-7} \cline{8-9}
\addlinespace[1mm]
Model & Method & Holistic & Analytic & & Holistic & Analytic & Holistic & Analytic \\
\hline
\addlinespace[1mm]
GPT4o & 5-shot &  25.58  & \textbf{74.42}  & & 23.81 & 76.19  & 59.10 & 40.90 \\
GPT4o (balanced) & 6-shot &  27.91  & 72.09  & & 19.05 & \textbf{80.95}  & 57.60 & \textbf{42.40} \\
\hline
\addlinespace[1mm]
%GPT4o-JP & 1-shot &   27.91  & 72.09  & & 23.81 & 76.19  & 54.45 & 45.55 \\
GPT4o-JP & 5-shot &   34.88  & 65.12  & & 38.10 & 61.90  & 63.86 & 36.14 \\
GPT4o-JP (balanced) & 6-shot &     \textbf{53.49}  & 46.51  & & \textbf{61.90} & 38.10  & \textbf{73.37} & 26.63 \\
\hline
\addlinespace[1mm]
%Gemma-3-12B \cite{team2025gemma} & 1-shot & 25.58  & 74.42  & & 19.05 & 80.95  & 44.40 &55.60  \\
Gemma-3-12B & 5-shot & 34.88  & 65.12 & & 19.05 & 80.95  & 54.10 & 45.90 \\
Gemma-3-12B (balanced) & 6-shot &  27.91  & \underline{\textbf{72.09}}  & & 14.29 &\underline{\textbf{85.71}}  & 54.80 & \underline{\textbf{45.20}} \\
\hline
\addlinespace[1mm]

VILA-JP-14B & 5-shot &  30.23  & \underline{\textbf{69.77}}  & & 28.57 & 71.43  & 55.10 & 44.90 \\
VILA-JP-14B (balanced) & 6-shot & \underline{\textbf{41.86}}  & 58.14  & & \underline{\textbf{33.33}} & 66.67  & \underline{\textbf{66.00}} & 34.00 \\
\hline
\addlinespace[1mm]
%VILA-JP-14B \cite{akiba2024evomodelmerge} & 1-shot &  13.95  & 86.05  & & 23.81 & 76.19  & 35.50 & 64.50 \\
Gemma-3-27B-JP & 5-shot &  27.91  & \underline{\textbf{72.09}} & & 28.57 & 71.43  & 65.20 & 34.80 \\
Gemma-3-27B-JP (balanced) & 6-shot &  39.53  & 60.47  & & \underline{\textbf{33.33}} & 66.67  & 64.60 & 35.40 \\
\hline
\end{tabular}
\vspace{-0.2cm}
\caption{Comparison analysis results examine different English and Japanese Vision-LLMs on Japanese cultural image benchmark datasets, including HERON-BENCH, JA-VLM, and our 1K JP-dataset. The results show that the \textit{Holistic} ratio increases in Japanese models and the \textit{Analytic} decreases in English models, when compared to the Western COCO dataset, suggesting that the dataset also contributes to certain cognitive styles.}
\label{tab:main-table-2}
\end{table*}

\section{Methodology}

\label{se:baseline}
Our framework consists of three main stages: image selection, caption generation, and textual classification via in-context learning (see Figure \ref{fig:enter-label}.)
\vspace{1pt}
\noindent{\textbf{Image Selection via Similarity Voting.}} 
To guarantee high confidence in object/background detection. We employ similarity voting using object class labels from four classifiers:  YOLOS \cite{DBLP:journals/corr/abs-2106-00666}, DEtection TRansformer (DETR) \cite{DBLP:journals/corr/abs-2005-12872}, Deformable DETR \cite{https://doi.org/10.48550/arxiv.2010.04159}, and  (DETA-Swin) \cite{ouyangzhang2022nms}. For each image, object labels are extracted, and pairwise cosine similarity is computed (6 combinations). If at least three out of six scores exceed a threshold $>$ 0.8, the image is accepted.
\vspace{1pt}

\noindent{\textbf{Caption Generation.}} Selected images are passed to vision language model for caption generation. We employ GPT-4o \cite{openai2024gpt4o} and Gemma-3-12 \cite{team2025gemma} for English caption generation, and for Japanese, GPT-4o-JP (GPT-4o with Japanese prompting), Gemma-3-27-JP, and a culturally-aware Japanese VILA-JP-14B model is used \cite{sasagawa2024constructing}.
\vspace{1pt}

\begin{table*}[ht]
    \centering
    \resizebox{\textwidth}{!}{
    \small
    \begin{tabular}{>{\centering\arraybackslash}m{0.3\textwidth} >{\raggedright\arraybackslash}m{0.7\textwidth}}
    \hline 
    Image & Image Caption \\
    \hline
        \addlinespace[1mm]
    \includegraphics[width=0.2\textwidth]{latex/images/COCO_val2014_000000018737.jpg}  & 
        \textbf{GPT4o:} The picture showcases \OF{a motorcycle} parked near a beach.  In the background, there is a sandy \BF{beach with sparse palm trees and a clear blue sky.} \OF{\WB{object first}}
    \newline
    \textbf{GPT-4o-JP:} \begin{CJK}{UTF8}{min}\BF{ビーチエリアの一部が写っており、砂地と背後にはいくつかのヤシの木が見えます}。写真の中央には赤と黒の\OF{オートバイ}が..。 \BF{\WB{background first}} \end{CJK}
    \newline
    \textbf{Translation:} This photograph shows part of a \BF{beach area}, with sand and several palm trees .. In the center of the photograph, \OF{a red and black motorcycle} is parked... 
    \\ 
    \hline
        \addlinespace[1mm]
    \includegraphics[width=0.2\textwidth]{latex/images/COCO_val2014_000000012192.jpg}  & 
   \textbf{GPT4o:} In the picture,  \OF{a young child} is playing by the \BF{shoreline of a beach.\OF{\WB{Object first}} }
    \newline 
\textbf{GPT-4o-JP:} \begin{CJK}{UTF8}{min}\BF{夕方または早朝に撮られたビーチ}の風景を写しています。\OF{手前にはラケットを持って遊んでいる子供が写っており} \BF{\WB{background first}}\end{CJK}   
    \newline
    \textbf{Translation:} This photograph depicts a \BF{beach scene taken either in the early morning} or in the evening. In the foreground, \OF{a child holding a racket} is seen playing.
    \\ 
    \hline
    \end{tabular}
    }
    \vspace{-0.2cm}
    \caption{Example images generated by GPT-4o with English and Japanese captions, highlighting differences in the order of background and object during image generation.}
    \label{tab:example_figure}
\end{table*}
\noindent{\textbf{Holistic/Analytic Classification.}} We analyze the generated caption using GPT-4o via In-Context Learning (ICL) \cite{brown2020language}. Specifically, we classify each caption into one of two categories: \textbf{Holistically [0]} and \textbf{Analytically [1]}. We use two prompting strategies: a 5-shot with five random examples (analytic/holistic) and a 6-shot, balanced examples with three examples from each class.

 \section{Experiments} 
We first validate our hypothesis on the 5K filtered COCO images using the captioning and ICL-based classification pipeline. Captions are generated in both English and Japanese. Each is classified as holistic or analytic.
We then extend the analysis to: (1) An additional 1K random COCO images (unfiltered) for findings validation, (2) Three Japanese-centric datasets (HERON, JA-VLM, and our 1K dataset), (3) Human evaluation on 200 randomly selected COCO images to compare with GPT-4o-JP, VILA-JP, and GPT-3o \cite{openai2025o3o4mini}.

Table~\ref{tab:main-table_paper} shows that for COCO images, GPT-4o-JP produces more holistic captions compared to VILA-JP, despite the latter being trained specifically on Japanese data. This suggests that language models may inherit biases from their pretraining setup rather than merely the language used. For English, VILA-JP produces results similar to GPT-4o and Gemma-12B. These findings suggest that these models exhibit an English bias when generating Japanese captions. 

When tested on culturally grounded Japanese image datasets, as in Table~\ref{tab:main-table-2}, the holistic ratio increases, indicating a strong influence of data content. For instance, GPT-4o-JP shows an increase in holistic classification from 26.10\% (COCO) to 73.37\% (1K-JP-dataset).
 The pattern holds across an additional evaluation on 1K random COCO images (Table~\ref{tab:1K random}) and aligns well with human annotations (Table~\ref{tab:human-eva}), affirming GPT-4o-JP’s tendency to generate holistic descriptions in Japanese. Table \ref{tab:example_figure} shows examples via GPT-4o on Japanese and English captions, reflecting different cognitive styles.

\begin{table}[t]
\resizebox{\columnwidth}{!}{%
\centering
\small
\begin{tabular}{lccc}
\hline
\addlinespace[1mm]
Model (6-shot) & Holistic & Analytic  \\
\addlinespace[1mm]
 
\addlinespace[1mm]
\hline
\addlinespace[1mm]
GPTV  \cite{achiam2023gpt} &  24.30  & 75.70  \\
GPT4o \cite{openai2024gpt4o} &   26.20 & 73.80 \\
LLAVA-Cot-11B \cite{xu2024llava} &   23.20 & 76.80 \\
%LlamaV-o1-11B \cite{thawakar2025llamav} & 0.0 & 0.0 \\
Gemma-3-12B \cite{team2025gemma} &   19.30 & 80.70 \\
Gemma-3-27B  &   19.20 & \textbf{80.80} \\

\addlinespace[1mm]
\hline 
\addlinespace[1mm]
%\st{EVO-VLM-JP-8B} \cite{akiba2024evomodelmerge} & 0.0  & 0.0  \\ 
VILA-JP-14B \cite{sasagawa2024constructing} &  24.00 & 76.00 \\ % \cite{chen2015microsoft} 
% Gemma-3-12B-JP  &   \textcolor{red}{00.00} & \textcolor{red}{00.00} \\
Gemma-3-27B-JP &   28.80 & 71.20 \\
%\st{Sarashina2-JP-14B} \cite{sarashina2-vision-14b} & 0.0 & 0.0  \\ 
GPT4o-JP  &   \textbf{35.50} & 64.50 \\
\hline
\end{tabular}
}

\vspace{-0.2cm}
   \caption{Comparison analysis results via different English and Japanese VLMs on random 1K images. }
    \label{tab:1K random}
\end{table}
\begin{table}[h!]
\resizebox{\columnwidth}{!}{%
\centering
\small
\begin{tabular}{lcccccccc}
\hline
\addlinespace[1mm]
 & \multicolumn{2}{c}{Human Eva \%} &  & \multicolumn{2}{c}{GPT4o 6-Shot \%} &  & \multicolumn{2}{c}{GPT-o3 \%} \\
\addlinespace[1mm]
\cline{2-3} \cline{5-6} \cline{8-9}
\addlinespace[1mm]
Model & H & A & & H & A & & H & A \\
\addlinespace[1mm]
\hline
\addlinespace[1mm]
VILA-JP & 15.33  & 84.67  & & 19.50 & 80.50 & & 19.00 & 81.00 \\ 
GPT4o-JP & 17.75  & 82.25  & & 29.00 & 71.00 & & 26.50 & 73.50 \\  
\addlinespace[1mm]
\hline
\end{tabular}
}\vspace{-0.2cm}
\caption{A human evaluation of 200 random COCO captions by three native Japanese speakers showed GPT-4o-JP tends to generate holistic Japanese descriptions.}

\label{tab:human-eva}
\end{table}

\begin{table}[t!]
%\resizebox{\columnwidth}{!}{%
\centering
\small
\begin{tabular}{lccc}
\hline
\addlinespace[1mm]
 &  \multicolumn{2}{c}{5K-6-Shot \%} \\
\addlinespace[1mm]
\cline{2-3} %\cline{5-6}
\addlinespace[1mm]
Datset & Holistic & Analytic  \\
\addlinespace[1mm]
 
\addlinespace[1mm]
\hline
\addlinespace[1mm]
%EVO VLM v2 & 14.50  & 85.50  & & 38.00 & 62.00  \\ 
Eng-original \cite{chen2015microsoft} &    12.34 \textcolor{white}{$\uparrow$} & 87.66 \textcolor{white}{$\uparrow$} \\ %\cite{chen2015microsoft} 
JP-MT &   27.54 \textcolor{gnuplot-blue}{$\uparrow$} & 72.46  \textcolor{gnuplot-red}{$\downarrow$}\\
Eng-BT &  15.20 \textcolor{gnuplot-blue}{$\uparrow$} & 84.80 \textcolor{gnuplot-red}{$\downarrow$} \\
\addlinespace[1mm]

\hline
\addlinespace[1mm]
JP-original  \cite{miyazaki2016}   & 27.48 \textcolor{white}{$\downarrow$} & 72.52 \textcolor{white}{$\uparrow$} \\ %cite{miyazaki2016}
ENG-MT &  16.26 \textcolor{gnuplot-red}{$\downarrow$} & 83.74 \textcolor{gnuplot-blue}{$\uparrow$} \\
JP-BT &  26.58 \textcolor{gnuplot-red}{$\downarrow$} & 73.42 \textcolor{gnuplot-blue}{$\uparrow$} \\
\hline
\end{tabular}

\vspace{-0.2cm}

      \caption{An experiment using human-written English COCO and human-translated Japanese COCO showed that Back-Translation (BT) improved the holistic score. In contrast, the Japanese version drops in holistic but rises in analytic, highlighting the impact of word order.}
    \label{tab:human-trans}
\end{table}

 \section{Discussion of Influencing Factors}
 \label{sec:analysis}

In this section, we examine more closely the factors that influence whether a VLM generates analytic or holistic descriptions.

\noindent{\bf{Model Size.}}
As shown in Table~\ref{tab:1K random}, larger models tend to better replicate culturally-influenced cognitive styles. For example, GPT-4o-JP and Gemma-27B-JP exhibit a more holistic descriptive tendency than the smaller VILA-JP-14B model. This suggests that increased model capacity may enable better generalization across cultural reasoning patterns, potentially due to richer representations and training exposure.

\noindent{\textbf{Language Structure.}}
Word order and linguistic structure also play a role. To explore this, we compare human-written English captions from the COCO-Caption dataset (original-English) \cite{chen2015microsoft} with their professionally translated Japanese counterparts from the Yahoo-COCO dataset (original-Japanese) \cite{miyazaki2016}. We then apply machine translation to perform a back-translation from Japanese to English using GPT-4o and the Google Translate API. Interestingly, we observe that the holistic/analytic classification ratio shifts by an average of 13.21\% after translation. Furthermore, similar shifts are observed when changing only the output language of the same model (\eg GPT-4o generating captions in English vs. Japanese). Back-translation into the original language tends to partially restore the initial descriptive pattern, with an average reversion of around 10\%. These results suggest that the structure and syntax of the output language itself contribute to shaping how visual scenes are described.

\noindent{\textbf{Cultural Context.}}
Finally, we find that the cultural origin embedded in datasets plays a major role. Images grounded in Japanese or East Asian contexts naturally elicit more holistic descriptions. For instance, in our Japanese cultural datasets, the proportion of holistic outputs reaches 73\% for GPT-4o-JP, an increase of 47\% points compared to COCO images. This reinforces the notion that the cultural characteristics embedded in the visual content and/or training data influence the descriptive tendencies of the models.

\section{Conclusion}
Our study shows that both the size of the model and the output language influence how vision-language models inherit cognitive style or cultural awareness. This effect increases when the dataset is culturally aligned with either East-Asian or Western perspectives, suggesting that both linguistic and cultural alignment are important for understanding culture-specific behaviors in large language models.

\section*{Limitation}
This study builds upon the work of \cite{Masuda2001}, which focuses exclusively on English and Japanese. However, in future work, we will explore different Asian languages and Latin languages to broaden the scope of this research. Although GPT-4o has a better tokenizer for the Japanese language, the limitations of high-capability Japanese VLMs restrict direct comparisons with the English language model.

\section*{Ethics statement}
This work aims to investigate whether Vision large language models (VLMs), either trained from scratch or fine-tuned on culturally diverse data, inherit human-like cognitive differences in holistic and analytical thinking that are known to vary across cultures, particularly between Western and East Asian contexts. However, understanding whether such culturally influenced thinking patterns are reflected in VLM behavior is not only a technical inquiry but also one that requires cultural sensitivity and awareness. It is essential to ensure that representations of different cultural cognitive styles are accurate, fair, and free from reductive biases or stereotypes.

\section*{Acknowledgment}

This work has received funding from the EU H2020 program under the SoBigData++ project (grant agreement No. 871042), by the CHIST-ERA grant No. CHIST-ERA-19-XAI-010, (ETAg grant No. SLTAT21096), and partially funded by  HAMISON project.

\bibliography{custom}

\appendix
\section*{Appendix}
\label{sec:appendix}

\section{Observation on Cognitive Style Classification}

The image caption that a model is going to output and its cognitive style classification depend on three factors. Most consistently on the language of output. Secondly, the style is better replicated by models that are bigger in size. And thirdly, datasets that are biased towards a certain cognitive style will leak these biases into the captions generated by the models.
In this paper, we demonstrate that models, such as GPT-4o, which are trained on large amounts of data, inherit linguistic and cultural behaviors, including language structure, that align with human cognition.  Specifically, when comparing English and Japanese VLM, there is a different ratio in holistic and analytic descriptions. That aligned with the same category as observed in humans, that Japanese VLMs are more likely than English VLMs to start with the background first than an object. 

\section{Other Western Languages}
Although the work by \cite{Masuda2001} focuses on English as the most widely spoken Western language, we also investigate another language, specifically Latin-based Spanish. As shown in Table \ref{tab:diff-lang}, the results align with our findings in English in terms of the cognitive ratio. That is also evident in the 1K Japanese dataset, which shows the same ratio shift in patterns toward Holistic and Analytic styles, with 67.60\% holistic and 32.40\% analytic. This confirms that the dataset is also a factor influencing a preference for a certain cognitive style.

\begin{table}[t!]
\resizebox{\columnwidth}{!}{%
\centering
\small
\begin{tabular}{lcccccccc}
\hline
\addlinespace[1mm]
 & \multicolumn{2}{c}{1K \%} &  & \multicolumn{2}{c}{2K \%} &  & \multicolumn{2}{c}{5K \%} \\
\addlinespace[1mm]
\cline{2-3} \cline{5-6} \cline{8-9}
\addlinespace[1mm]
Model & H & A & & H & A & & H & A \\
\addlinespace[1mm]
\hline
\addlinespace[1mm]
GPT4o-EN & 20.80  & 79.20  & & 21.05 & 78.95 & & 19.80 & 80.20 \\ 
GPT4o-ES & 18.10  & 81.90  & & 19.35 & 80.65 & & 19.12 & \textbf{80.88} \\ 
GPT4o-JP & 26.10  & 73.90  & & 27.75 & 72.25 & & \textbf{28.64} & 71.36 \\  
\addlinespace[1mm]
\hline
\end{tabular}
}\vspace{-0.2cm}
\caption{A Spanish-language experiment shows GPT-4o outputting similar analytic captions with the same cognitive ratio as its English counterpart.}
\label{tab:diff-lang}
\end{table}

\section{Additional Information on Experiments}

\noindent{\textbf{Japanese Dataset.}} We also introduce a 1K dataset of images captured in Japan. It was distilled from 6,435 images captured by The Pioneer\footnote{\url{https://huggingface.co/datasets/ThePioneer/japanese-photos}}. The process of cleaning the dataset included using GPT-4o-mini as a judge to remove images that included screens, digital displays, presentations, posters, informational signs, or banners. After that, a CNN-based deduplication algorithm\footnote{(\url{https://github.com/idealo/imagededup}}  was used to remove duplicates, reducing the size to 3759 images. Out of these, 1K were hand-picked with the same ratio of background- or object-prominent scenes.

\noindent{\textbf{Implementation details.}} 
We outline the implementation details of the models used in the experiments.  All experiments were conducted using a two V100 GPU with 32 GB VRAM. The GPT-4V (appendix) \texttt{gpt-4-1106-vision-preview} and GPT-4o \texttt{gpt-4o-2024-08-06} models used the default temperature for generations.

\noindent{\textbf{Additional Results.}} Table~\ref{tab:main-table-1-full} presents additional results on the COCO dataset. Table~\ref{tab:main-table-2-full} shows additional results on the culture-aware Japanese dataset. Table \ref{tab:my_label_v1} and Table \ref{tab:my_label_v2} show examples of images with their English and Japanese captions, showing the differences in order of background and object.

\begin{table*}[ht]
\centering
\small
 \resizebox{0.9\textwidth}{!}{
\begin{tabular}{l c c c c c c c c c}
\hline
 &  & \multicolumn{2}{c}{1K \%} &  & \multicolumn{2}{c}{2K \%} &  & \multicolumn{2}{c}{5K \%} \\
\addlinespace[1mm]
\cline{3-4} \cline{6-7} \cline{9-10}
\addlinespace[1mm]
Model & Method & Holistic & Analytic & & Holistic & Analytic & & Holistic & Analytic \\
\hline

%\hline

\addlinespace[1mm]

Human   & 5-shot &  10.10 & 89.90  && 10.80 & 89.20 & & 10.70 & 89.30 \\ 
Human (balanced)  & 6-shot &  10.60 & 89.40  && 13.10 & 86.90 & & 13.10 & 86.90 \\ 

\addlinespace[1mm]
\hline
\addlinespace[1mm]
GPT4V   & 5-shot &  19.12 & 80.88  && 17.11 & 82.89 & & 15.95 & \textbf{84.05} \\ 
GPT4V (balanced)  & 6-shot &  20.12 & 79.88  && 19.26 & 80.74 & & 18.43 & 81.57 \\ 
\addlinespace[1mm]
\hline
\addlinespace[1mm]

GPT4o   \cite{openai2024gpt4o} & 5-shot &  22.60 & 77.40  && 20.25 & 79.75 & & 19.22 & 80.78 \\ 
GPT4o (balanced)  & 6-shot &  20.80 & 79.20  && 21.05 & 78.95 & & 19.80 & 80.20 \\                                             
\addlinespace[1mm]
\hline
\addlinespace[1mm]

Gemma-3-12B  \cite{team2025gemma} & 5-shot & 13.40 & 86.60  && 13.30 & 86.70 & & 14.16 & 85.84 \\ 
Gemma-3-12B (balanced)  & 6-shot &  13.70 & 86.30  && 13.15 & 86.85 & & 13.78 & 86.22 \\        
\addlinespace[1mm]
\hline
\addlinespace[1mm]
Gemma-3-27B  & 5-shot & 14.80 & 85.20  && 14.40 & 85.60 & & 14.50 & 85.50 \\ 
Gemma-3-27B (balanced)  & 6-shot &  13.00 & 87.00  && 13.00 & 87.00 & & 13.54 & 86.46 \\           
\addlinespace[1mm]
\hline
 \addlinespace[1mm]
VILA-JP-14B  \cite{sasagawa2024constructing} &  5-shot    &   11.60 & 88.40  &&  12.80 & 87.20 & &  13.88  &  86.12  \\ 
VILA-JP-14B (balanced)  &  6-shot    &   14.70 & 85.30  && 15.75  & 84.25 & &  16.90  &  83.10   \\  \addlinespace[1mm]
 \hline
\addlinespace[1mm]
GPT4o-JP   & 5-shot &   24.60 & 75.40  &&  24.15 & 75.85 & &  23.64 & 76.36 \\ 
GPT4o-JP (balanced)  & 6-shot &  26.10 & 73.90  && 27.75 & 72.25 & & \textbf{28.64} & 71.36 \\ 
\addlinespace[1mm]
\hline
\addlinespace[1mm]
Gemma-3-27B-JP  & 5-shot & \textcolor{black}{13.00} & \textcolor{black}{87.00}  && \textcolor{black}{13.00} & \textcolor{black}{87.00} & &\textcolor{black}{13.50}  & \underline{\textbf{86.50}} \\ 
Gemma-3-27B-JP (balanced)  & 6-shot &  \textcolor{black}{19.70} & \textcolor{black}{80.30}  && \textcolor{black}{19.55} & \textcolor{black}{80.45} & & \textcolor{black}{\underline{\textbf{21.14}}} & \textcolor{black}{78.86} \\           
\addlinespace[1mm]
\hline

%\hline
%\addlinespace[1mm]
%\hline
\end{tabular}}
\vspace{-0.2cm}
\caption{\textbf{FULL result: COCO dataset}. Comparison analysis results via different English and Japanese Vision-LLMs on selecting images from COCO dataset, with GPT-4o serving as the judge of the model selecting image captioning description starting with backgrounds \textit{ holistic} or object \textit{Analytic}. The results indicate that larger models, such as GPT-4o, better mimic similar human cognitive styles, especially for Japanese \textit{Holistic}. \textbf{Bold} indicates the highest score among large models, while \underline{\textbf{Underline}} indicates the highest score among smaller models.}

  \label{tab:main-table-1-full}

\end{table*}

\begin{table*}[ht]
\centering
\small
\begin{tabular}{lcccccccc}
\hline
 &  & \multicolumn{2}{c}{HERON-BENCH \%} &  & \multicolumn{2}{c}{JA-VLM \%} & \multicolumn{2}{c}{1K- JP-dataset  \%} \\
\addlinespace[1mm]
\cline{3-4} \cline{6-7} \cline{8-9}
\addlinespace[1mm]
Model & Method & Holistic & Analytic & & Holistic & Analytic & Holistic & Analytic \\
\hline
\addlinespace[1mm]
GPT4o & 5-shot &  25.58  & \textbf{74.42}  & & 23.81 & 76.19  & 59.10 & 40.90 \\
GPT4o (balanced) & 6-shot &  27.91  & 72.09  & & 19.05 & \textbf{80.95}  & 57.60 & \textbf{42.40} \\
\hline
\addlinespace[1mm]
GPT4o-JP & 5-shot &   34.88  & 65.12  & & 38.10 & 61.90  & 63.86 & 36.14 \\
GPT4o-JP (balanced) & 6-shot &     \textbf{53.49}  & 46.51  & & \textbf{61.90} & 38.10  & \textbf{73.37} & 26.63 \\
\hline
\addlinespace[1mm]
Gemma-3-12B & 5-shot & 34.88  & 65.12 & & 19.05 & 80.95  & 54.10 & \textbf{\underline{45.90}} \\
Gemma-3-12B (balanced) & 6-shot &  27.91  & 72.09  & & 14.29 &\textbf{\underline{85.71}}  & 54.80 & 45.20 \\
\hline
\addlinespace[1mm]
Gemma-3-27B & 5-shot &  27.91  & 72.09 & & 23.81 & 76.19  & 55.30 & 44.70 \\
Gemma-3-27B (balanced) & 6-shot &  25.58  & \textbf{\underline{74.42}}  & & 23.81 & 76.19  & 54.90 & 45.10 \\
%\hline
%\addlinespace[1mm]
\hline
\addlinespace[1mm]
Gemma-3-27B-JP & 5-shot &  27.91  & 72.09 & & 28.57 & 71.43  & 65.20 & 34.80 \\
Gemma-3-27B-JP (balanced) & 6-shot &  39.53  & 60.47  & & \textbf{\underline{33.33}} & 66.67  & 64.60 & 35.40 \\
\hline
\addlinespace[1mm]
VILA-JP-14B & 5-shot &  30.23  & 69.77  & & 28.57 & 71.43  & 55.10 & 44.90 \\
VILA-JP-14B (balanced) & 6-shot &  \textbf{\underline{41.86}}  & 58.14  & & \textbf{\underline{33.33}} & 66.67  & \textbf{\underline{66.00}} & 34.00 \\
\hline
\end{tabular}
\vspace{-0.2cm}
\caption{\textbf{FULL result: Culture-Aware Japanese dataset}. Comparison analysis results examine different English and Japanese Vision-LLMs in Japanese cultural image benchmark datasets, including HERON-BENCH, JA-VLM, and JP-dataset, with GPT-4o serving as the judge for determining whether the model's image captioning description starts with backgrounds \textit{holistic} or objects \textit{analytic}. The results indicate that the holistic ratio increases in Japanese models compared to the Western COCO dataset, suggesting that the dataset also contributes to certain cognitive styles. \textbf{Bold} indicates the highest score among large models, while \underline{\textbf{Underline}} indicates the highest score among smaller models.}
\label{tab:main-table-2-full}
\end{table*}

\begin{table*}[ht]
    \centering
    \resizebox{\textwidth}{!}{
    \small
    \begin{tabular}{>{\centering\arraybackslash}m{0.3\textwidth} >{\raggedright\arraybackslash}m{0.7\textwidth}}
    \hline 
    Image & Image Caption \\
    \hline
        \addlinespace[1mm]

    % 1
    \includegraphics[width=0.2\textwidth]{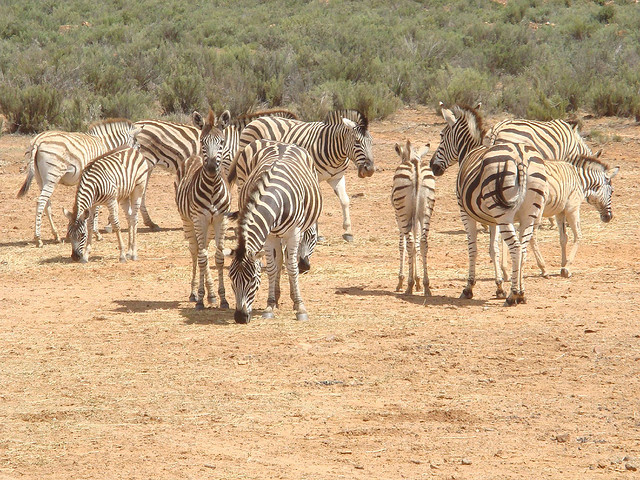}  & 
    \textbf{GPT4o:} The picture shows a \OF{group of zebras} gathered in a semi-arid environment. The zebras are \BF{standing on dry, gravelly ground} \OF{\WB{object first}}
    \newline
    \textbf{GPT-4o-JP:} \begin{CJK}{UTF8}{min}この写真には、\OF{複数のシマウマ (several zebras)} が草原のような場所に集まり草を食べている様子..。\BF{背景には低い緑の茂み (low green bushes)} が広がっており (略) \OF{\WB{object first}} \end{CJK}
    \newline
    \textbf{VILA-JP:} \begin{CJK}{UTF8}{min} 画像には、\BF{乾燥した土のフィールド (a dry soil field)} で草を食べている\OF{シマウマの群れ (a herd of zebras)} が写っています。 \BF{\WB{background first}}\end{CJK}
    \\ 
      \addlinespace[1mm]
    \hline
        \addlinespace[1mm]

    %2
    \includegraphics[width=0.2\textwidth]{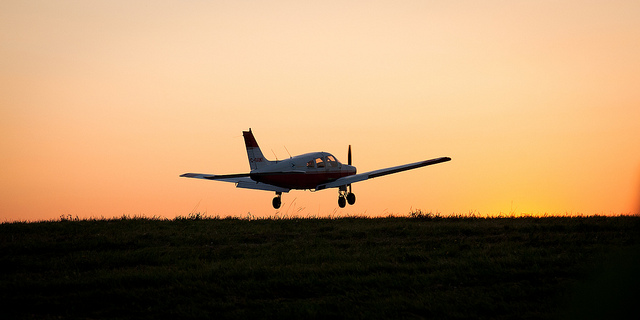}  & 
    \textbf{GPT4o:} The picture depicts a small \OF{airplane}, likely a single-engine aircraft, in the process of landing.. The \BF{sky} is filled with warm hues of orange and yellow..\OF{\WB{object first}}
    \newline 
    \textbf{GPT-4o-JP:} \begin{CJK}{UTF8}{min}この写真には、夕暮れ時の地平線上に\OF{小型飛行機 (a small propeller plane)} が写っています。 背景には、\BF{オレンジ色やピンク色の美しい夕焼 (a beautiful orange and pink sunset)} (略) \OF{\WB{object first}} \end{CJK}   
    \newline
    \textbf{VILA-JP:} \begin{CJK}{UTF8}{min}画像には、\BF{夕焼けの空を背景にした (sunset sky as background)} 小型のプロペラ機が写っています。この\OF{飛行機 (a small propeller plane)} は草地の上を低く飛行 (略) \BF{\WB{background first}}\end{CJK}
    \\ 
      \addlinespace[1mm]
    \hline
        \addlinespace[1mm]

     % 3
  \includegraphics[width=0.2\textwidth]{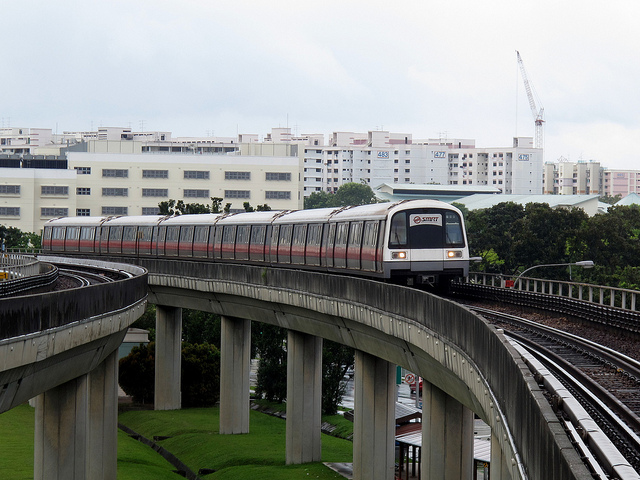}  & 
    \textbf{GPT4o:} The picture depicts a \OF{modern train} running on elevated railway tracks... Below the tracks is a grassy area, and \BF{in the background, there are several large buildings}, possibly residential or commercial, as well as a construction crane. \OF{\WB{object first}}
    \newline
    \textbf{GPT-4o-JP:} \begin{CJK}{UTF8}{min} この写真には、高架を走行する\OF{電車 (train)} が写っています。 (略) \BF{背景にはいくつかの建物 (buildings)} が見え、(略) \OF{\WB{object first}} \end{CJK}
    \newline
    \textbf{VILA-JP:} \begin{CJK}{UTF8}{min} 画像には、都市の上空を走る\OF{列車 (train)} が写っています。この列車は高架の線路を走行しており、\BF{周囲にはいくつかの建物 (buildings)} が見えます。 (略) 全体的に、都市の活気ある風景が広がっています。 \OF{\WB{object first}}\end{CJK}
    \\ 
      \addlinespace[1mm]
    \hline
        \addlinespace[1mm]

    % 4
   \includegraphics[width=0.2\textwidth]{latex/images/COCO_val2014_000000026363.jpg}  & 
    \textbf{GPT4o:} This photograph shows several \OF{elephants} being used for carrying tourists in a location that appears to be an \BF{ancient or historical site}.. \OF{\WB{object first}}
    \newline
    \textbf{GPT-4o-JP:} \begin{CJK}{UTF8}{min}この写真には、\BF{アユタヤの遺跡 (ruins of Ayutthaya) が背景に}映し出されており、\OF{複数のゾウ (elephants)} が見えます。(略)
    \BF{\WB{background first}} \end{CJK}
    \newline
    \textbf{VILA-JP:} \begin{CJK}{UTF8}{min} この画像には、\OF{数頭の象 (elephants)} が描かれており、その上には人々が乗っています。(略)
    象の周りには\BF{緑豊かな風景 (lush landscape)} が広がっており、自然の中での楽しいひとときを感じさせます。(略)
    \BF{\WB{background first}}\end{CJK}
    \\ 
      \addlinespace[1mm]
    \hline
        \addlinespace[1mm]

    % 5
   \includegraphics[width=0.2\textwidth]{latex/images/COCO_val2014_000000003255.jpg}  & 
    \textbf{GPT4o:} The picture shows a \OF{group of people} standing on a \BF{snowy landscape} near a mountain. The mountain is covered with snow... \OF{\WB{object first}}
    \newline
    \textbf{GPT-4o-JP:} \begin{CJK}{UTF8}{min}この写真は、\BF{美しい雪景色が広がる山岳地帯 (mountainous area with a beautiful snowy landscape)} を写しています。背景には、雪で覆われた険しい岩山がそびえ立ち、青空が広がっています。前景には\OF{一群の人々 (group of people)} がおり、(略)
    \BF{\WB{background first}} \end{CJK}
    \newline
    \textbf{VILA-JP:} \begin{CJK}{UTF8}{min} 画像には、\BF{雪に覆われた山 (snow covered mountain)} の斜面でスキーを楽しむ\OF{人々のグループ (group of people)} が写っています。彼らはスキーを履いており、雪の上を滑ったり、立ち止まったりしています。(略) \BF{\WB{background first}}\end{CJK}
    \\ 
      \addlinespace[1mm]
    \hline
        \addlinespace[1mm]

    % 6
   \includegraphics[width=0.2\textwidth]{latex/images/COCO_val2014_000000007816.jpg}  & 
    \textbf{GPT4o:} The picture shows a \OF{person} riding a sports motorcycle on a curvy road. The rider is wearing...  In the \BF{background, there is a grassy area with a fence lining the road}. A group of people are standing and sitting by the fence... \OF{\WB{object first}}
    \newline
    \textbf{GPT-4o-JP:} \begin{CJK}{UTF8}{min} 写真は次のようなシーンを描写しています：  \BF{緑の牧草地 (green pastures)} に囲まれた舗装道路上で、\OF{オートバイに乗ったライダー (a rider on a motorcycle)} がカーブを曲がっています。ライダーは白いヘルメットと白いジャケットを着用しており、オートバイも白いボディカラーをしています。(略) \BF{\WB{background first}} \end{CJK}
    \newline
    \textbf{VILA-JP:} \begin{CJK}{UTF8}{min} 画像には、\BF{緑豊かな丘 (lush hill)} の近くの道路を走る\OF{バイクに乗った男性}が写っています。彼は白いヘルメットをかぶり、バイクを巧みに操っています。周囲には他の人々も見え、彼らはバイクの走行を見守っているようです。  バイクは道路の中央を走っており、その周りには数人の観客が立っています。(略) \BF{\WB{background first}}\end{CJK}
    \\ 
      \addlinespace[1mm]
    \hline
        \addlinespace[1mm]
    \end{tabular}
    }
    \vspace{-0.2cm}
    \caption{Example of images with their English and Japanese captions showing the differences in order of background and object.}
    \label{tab:my_label_v1}
\end{table*}

\begin{table*}[ht]
    \centering
    \resizebox{\textwidth}{!}{
    \small
    \begin{tabular}{>{\centering\arraybackslash}m{0.3\textwidth} >{\raggedright\arraybackslash}m{0.7\textwidth}}
    \hline 
    Image & Image Caption \\
    \hline
        \addlinespace[1mm]

    % 7
    \includegraphics[width=0.2\textwidth]{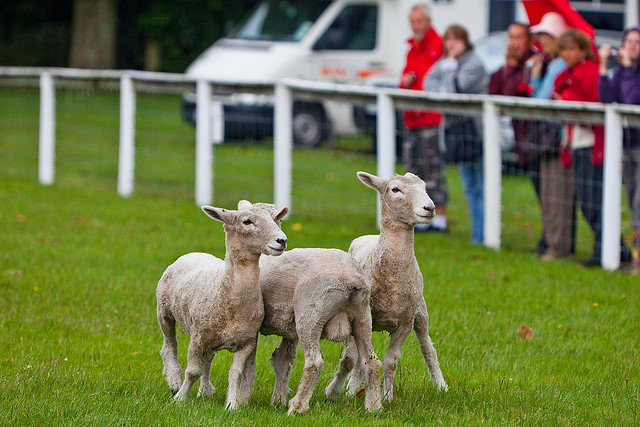}  & 
    \textbf{GPT4o:} The picture shows \OF{three sheep} standing close together in a grassy area, likely a field or a park. There is a white fence separating the sheep from \BF{a group of people in the background}. Some of the people appear to be watching the sheep, ...
    \OF{\WB{object first}}
    \newline
    \textbf{GPT-4o-JP:} \begin{CJK}{UTF8}{min}この写真には、芝生の上に\OF{3匹の羊 (three sheep)} が写っています。羊たちは近くに集まって立っています。\BF{背景には白いフェンスがあり、その向こうに人々 (people)}が集まって羊を見ています。(略)
    \OF{\WB{object first}} \end{CJK}
    \newline
    \textbf{VILA-JP:} \begin{CJK}{UTF8}{min}\OF{羊 (sheep)}の近くには、\BF{いくつかのハンドバッグ (handbag) が見え、人々 (people) がそれを持っている様子}が伺えます。また、\BF{背景にはトラック (truck)} が停まっており、羊たちの存在を引き立てています。人々は羊を観察したり、写真を撮ったりしているようで、賑やかな雰囲気が漂っています。   \OF{\WB{object first}} \end{CJK}
     \\ 
    \addlinespace[1mm]
    \hline
    \addlinespace[1mm]

    % 8
    \includegraphics[width=0.2\textwidth]{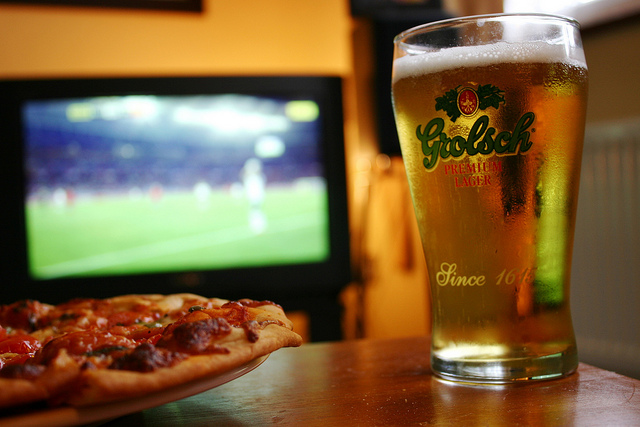}  & 
    \textbf{GPT4o:} The foreground shows a close-up of  \OF{a glass of beer}, branded "Grolsch Premium Lager," which is a well-known beer.  In the background, there is a \BF{television screen displaying what looks like a sports game, possibly soccer (football)} ...
    \OF{\WB{object first}}
    \newline 
    \textbf{GPT4o-JP:} \begin{CJK}{UTF8}{min}この写真には、\BF{サッカーの試合を映しているテレビ画面 (soccer match in TV)} と、手前にはピザとビールのグラスが写っています。グラスには「Grolsch Premium Lager」と書かれており、ビールが注がれています。 (略)
    \BF{\WB{backround first}}\end{CJK}   
    \newline
    \textbf{VILA-JP:} \begin{CJK}{UTF8}{min}この画像には、テーブルの上に\OF{ビールのグラスとピザ (beer glass and pizza)} が置かれています。ビールのグラスはテーブルの中央に位置し、その隣にはピザが皿に盛られています。テーブルの奥にはテレビがあり、スポーツの試合が放送されている様子が伺えます。 (略) 
    \OF{\WB{object first}}\end{CJK}   
    \\ 
    \addlinespace[1mm]
    \hline
    \addlinespace[1mm]
    % 7
    \includegraphics[width=0.2\textwidth]{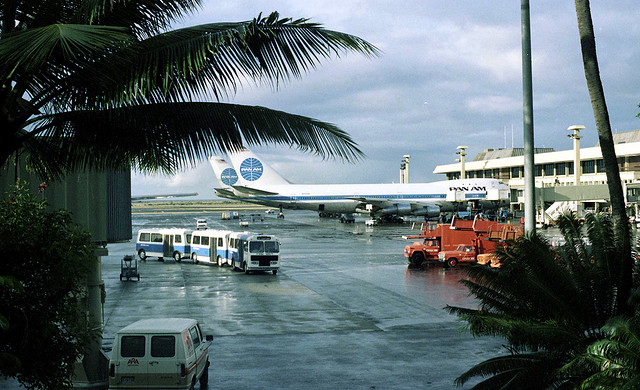}  & 
    \textbf{GPT4o:} This picture depicts a \BF{scene at an airport}. The image captures \OF{a Pan Am airplane} on the tarmac, with its distinct blue globe logo visible on the tail...
    \BF{\WB{backround first}}
    \newline
    \textbf{GPT-4o-JP:} \begin{CJK}{UTF8}{min}この写真は、\BF{空港のターミナル (terminal builds)} と駐機場の風景を描いています。前景には、\OF{パームツリー (palm trees) や植物が見え、その背後にはいくつかの車両 (vehicles) が駐車されています}。(略)
    背景には、空港ターミナルビルが並んでおり、建物の屋根には複数の照明塔があります。
    \BF{\WB{background first}} \end{CJK}
    \newline
    \textbf{VILA-JP:} \begin{CJK}{UTF8}{min}画像には、空港の滑走路に停まっている\OF{大型の旅客機 (airplane)} が写っています。この飛行機は、空港のターミナルに接続されたゲートに駐機しており、周囲にはいくつかのトラックが見えます。  
    全体的に、\BF{空港の活気ある雰囲気 (lively atmosphere of the airport)} が感じられるシーンです。
    \OF{\WB{object first}} \end{CJK}
    \newline
    \\ 
      \addlinespace[1mm]
    \hline
        \addlinespace[1mm]

    %8
    \includegraphics[width=0.2\textwidth]{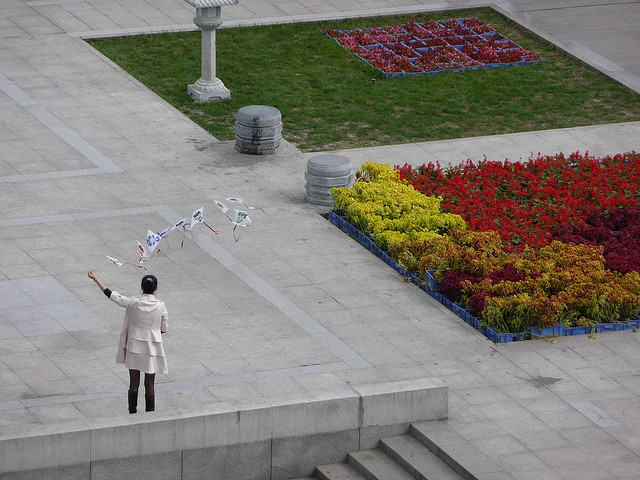}  & 
    \textbf{GPT4o:} The picture shows an aerial view of \OF{a person flying a kite} in an open, paved area. 
    The person is dressed in a light-colored coat and black knee-high boots and is standing near some small, \BF{colorful flower beds}. ...
    \OF{\WB{object first}}
    \newline 
    \textbf{GPT-4o-JP:} \begin{CJK}{UTF8}{min}写真には、\BF{広々とした石畳 (wide stone-paved area)} の場所に一人の人物 (a person) が写っています。(略)
    \BF{\WB{background first}}  \end{CJK}
    \newline
    \textbf{VILA-JP:} \begin{CJK}{UTF8}{min}画像には、\BF{美しい花が咲いている庭}の近くで、\OF{女性が凧を飛ばしている (a woman is flying a kite)} 様子が写っています。(略)
    \BF{\WB{background first}} \end{CJK} 
    \\ 
      \addlinespace[1mm]
    \hline
        \addlinespace[1mm]

     % 9
    \includegraphics[width=0.2\textwidth]{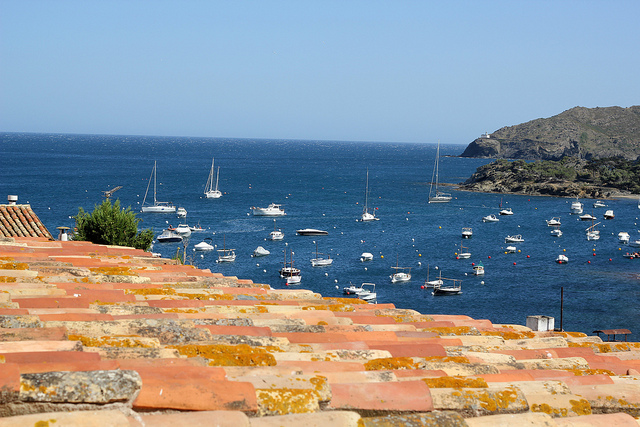}  & 
    \textbf{GPT4o:} The picture shows a \BF{picturesque coastal scene}. In the foreground, there is \OF{a roof covered with terracotta tiles}, giving a warm, rustic feel to the image.
    \BF{\WB{background first}}
    \newline 
    \textbf{GPT-4o-JP:} \begin{CJK}{UTF8}{min}この写真には、\BF{美しい海の風景 (beautiful sea scenery)}が映っています。前景にはオレンジ色や赤茶色の瓦屋根が見え、その先には\BF{青い海 (blue sea)} が広がっています。(略)
   \BF{\WB{background first}} \end{CJK}
    \newline
    \textbf{VILA-JP:} \begin{CJK}{UTF8}{min}この画像には、海の近くにある\OF{建物の屋根 (building roof)} が写っています。屋根は赤とベージュの色合いで、\BF{周囲には多くのボート (many boats around)} が浮かんでいます。
    \OF{\WB{object first}} \end{CJK} 
    \\ 
      \addlinespace[1mm]
    \hline
        \addlinespace[1mm]

    % 10
    \includegraphics[width=0.2\textwidth]{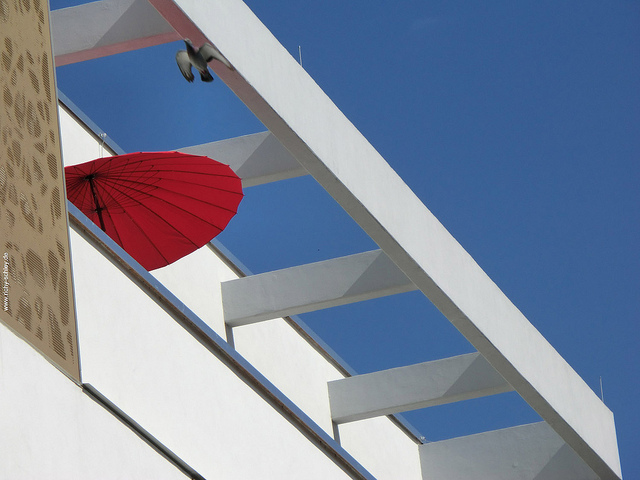}  & 
    \textbf{GPT4o:} This picture features a striking \BF{architectural scene set against a clear blue sky}...
    %The structure in the photo is composed of clean, white lines and geometric shapes} that create a modern aesthetic... 
    % Several horizontal beams extend from the building, adding to the geometric design.
    Near the center of the image, a \OF{red umbrella} stands out, providing a vivid contrast to the white and blue background... 
    % The umbrella's vibrant color draws the viewer's attention and adds an element of interest to the minimalist composition. Additionally, there is a sculptural element resembling bird wings near the top of the image, contributing a dynamic touch to the scene.  Overall, the picture captures a harmonious balance of color, shape, and form, highlighting the beauty of simplicity in architectural design.
    \BF{\WB{background first}}
    \newline 
    \textbf{GPT-4o-JP:} \begin{CJK}{UTF8}{min}この写真には、\BF{青空を背景 (blue sky background)}に\OF{白い建物 (white building)}が描かれています。(略)
    % 建物の構造は直線的で、白い壁と水平に伸びる白い梁が特徴です。
    % 建物の一部には鮮やかな赤の傘が見え、階段かバルコニーのエリアに配置されているように見えます。傘の存在が写真全体に鮮やかなアクセントをもたらしています。また、写真の上部には鳩が飛んでいる様子も写っています。全体としてモダンでミニマリズムな印象の写真です。 
    %\OF{\WB{object first}} 
      \BF{\WB{background first}}\end{CJK}
    \newline
    \textbf{VILA-JP:} \begin{CJK}{UTF8}{min}画像には、\OF{白い建物の外に赤い傘 (red umbrella outside a white building)} が立てかけられています。(略)
    % この傘は建物の側面に寄りかかっており、その鮮やかな色合いが目を引きます。傘の近くには、建物の構造の一部が見え、全体的にシンプルでモダンな印象を与えています。  傘の下には、建物の一部が見え、周囲には他の物体は見当たらず、傘が際立っています。このシーンは、静かな日常の一コマを切り取ったような雰囲気を醸し出しています。
    \OF{\WB{object first}} \end{CJK}   
    \\ 
      \addlinespace[1mm]
    \hline
        \addlinespace[1mm]

    \end{tabular}
    }
    \vspace{-0.2cm}
    \caption{Example of images with their English and Japanese captions showing the differences in order of background and object. }
    \label{tab:my_label_v2}
\end{table*}

\end{document}